\documentclass[a4paper,conference]{IEEEtran}

\ifCLASSOPTIONcompsoc
  \usepackage[nocompress]{cite}
\else
  \usepackage{cite}
\fi

\ifCLASSINFOpdf
  \usepackage[pdftex]{graphicx}
  \graphicspath{{./}{media/}}
\else
\fi

\usepackage{amsmath}
\usepackage{amssymb, bm}
\usepackage{color}
\usepackage{theorem}

\usepackage{graphicx}
\usepackage{multicol}
\usepackage{enumitem}
\usepackage{subfig}

\newtheorem{prp}{Proposition}

\newcommand{\sign}[1]{{\sigma(#1)}}
\begin{document}

\title{Pose estimation of a single circle  using
default intrinsic calibration}

\author{
	\IEEEauthorblockN{Damien \textsc{Mariyanayagam}, Pierre \textsc{Gurdjos}, \\Sylvie \textsc{Chambon} and Vincent \textsc{Charvillat}}
	\IEEEauthorblockA{IRIT (INP-ENSEEIHT), Toulouse, France\\
	Email: dmariyan@enseeiht.fr}
	\and
	\IEEEauthorblockN{Florent \sc{Brunet}}
	\IEEEauthorblockA{Ubleam, Toulouse, France\\
	Email: florent.brunet@ubleam.com
	}
}
\maketitle

\begin{abstract}
Circular markers are planar markers which offer great performances for detection
and pose estimation. For an uncalibrated camera with an unknown focal  length,
at least the images of at least two coplanar circles are generally required
to recover
their poses. Unfortunately, detecting more than one ellipse in the image
must be tricky and time-consuming, especially regarding concentric circles.
On the other hand, when the camera is calibrated, one circle suffices but
the solution is twofold and can hardly be disambiguated. 
 Our contribution is
to put beyond this
limit by dealing with the  uncalibrated case of a  camera seeing one circle
and discussing how to remove the ambiguity.
We propose a new problem formulation that enables to show how to detect geometric
configurations in which the ambiguity can be removed.
Furthermore, we introduce the notion of default camera intrinsics and show,
using intensive empirical
works, the surprising observation that  very approximate calibration can
lead to accurate circle pose
estimation.\end{abstract}

%
\IEEEpeerreviewmaketitle




\section{Introduction}


The problem of estimating the pose of a camera  (or dually  of a 3D object)
from a set of 2D  projections in a single view has been widely studied in
the computer vision literature for a long time~\cite{Szeliski2011}.
The  ``minimal'' problem i.e., which requires the minimal
amount of information necessary, is known as the perspective-3-point-problem
(P3P) and consists in recovering
the pose of a calibrated camera from three 3D-2D point correspondences.
Many solutions are available for the general case when more information is
available.  When the environment in the scene can be controlled, artificial
features with known positions are very often deployed in the scene. They
are used
in a wide range of applications, especially when a reliable
reference is needed to, e.g., in cluttered or textureless environments. The
most popular 
artificial features are probably  coplanar features\cite{Fiala2010} whose
layout in\ 2D space defines a so-called 
planar marker.
The mapping between a planar marker and its image is a 2D projective transformation
known as (world-to-image) homography and can be estimated 
from at least four world-to-image correspondences (the most simple planar
marker is a square).
Once the camera is calibrated, the decomposition of the  homography matrix
allows to recover the pose of the camera (or dually that of the plane). 

Other well-known artificial markers that have been recently investigated again are those consisting
of coplanar circles~\cite{Wu2004, Gurdjos2006, Kim2005, Bergamasco2011}.
The knowledge of at least two circle images (without any information on
their parameters on the support plane) allows to
compute a world-to-image homography without  ambiguity for all the spatial 
configurations of two circles except one~\cite{Gurdjos2006}.

Given a single circle image, it is well-known that a twofold solution exists
for the normal to the support plane (and so for the pose) \textit{only if}
the camera is calibrated \cite{Wu2004}. In this work, our contribution is to put beyond this
limit by dealing with the  case of an uncalibrated camera seeing one circle.
Actually, our starting point came from the surprising observation, learned from empirical
works, that  very approximate calibration can lead to accurate circle pose
estimation. Our idea is to use default
intrinsics  by designing
a generic camera model delivering a default focal length based on off-line
calibration of several smartphone cameras.

Our first contribution is to run
extensive experiments that assess how the inaccuracy of the calibration impacts the quality of the pose estimation. We found out that exact calibration may not be required
as small variations on the focal length does not affect the reprojection error of other reference coplanar points especially when the marker is far from the camera.
Our second contribution is to provide a new geometrical framework to state
the pose problem in which the issue of how to remove the twofold ambiguity
can be thoroughly investigated.

We review the related works in section~\ref{sec:related_work}.
Then in section~\ref{sec:reminder}, we remind the problem of recovering the pose from the  projection of a circle before introducing the solution
proposed in section~\ref{sec:vanish_line}.
The idea is to introduce a new way of computing the vanishing line
(dual to the plane normal) from one circle image.
Thanks to it, as the general method leads to two possible solutions, we show how under
some assumptions about the geometric configuration we can recover the correct
one.
Then, as we suppose that we work with uncalibrated images, we explain how
we select parameter values to obtain what we called default camera intrinsic parameters.
Finally in section~\ref{sec:experiment}, we evaluate our method in the context
of augmented reality.

\section{Related Work}
\label{sec:related_work}
A lot of existing works suggest to use a
set of features encoded in a planar pattern to simplify the pose estimation. Fiala et al. introduced a fiducial system~\cite{Fiala2010} and proposed the case of a special planar square marker.
Recent efficient algorithms allow to detect ellipses precisely and, in consequence, circles become features worth of interest.
The four projective and affine parameters of the world-to-image homography  (the remaining four define a similarity on the world plane \cite[p42]{RichardHartley2003}) can be
recovered by detecting the images of two special point of the support plane, known as circular points (e.g., see \cite[p52-53]{RichardHartley2003}) which are common to all circles.
Gurdjos et al.~\cite{Gurdjos2006} relied on the notion of pencil of circle images to formulate the problem of detecting the images of the circular points as a problem of intersection of lines, obtained from the degenerate members of the  pencil.
Kim et al.~\cite{Kim2005} proposed algebraic and geometric solutions in the case of concentric circles.
Calvet et al.~\cite{Calvet2016} described a whole fiducial
system using concentric circles which
allows to accurately detect the position of the image of the
circles’ common center under highly challenging
conditions. In a same vein,
Huang et al.~\cite{Huang2015} proposed to use the common self-polar triangle of concentric circles.

When using circular markers it is also possible to simplify the model of the camera to only depend on a sole focal length parameter.
Chen et al.~\cite{Chen2004a} autocalibrate the focal length using two or more coplanar circles.
The problem to solve contains two parameters: one ellipse and the focal length.
Then, two correspondences between a circle and an ellipse are necessary to estimate the focal length.
Based on the same method, Bergamasco et al.\cite{Bergamasco2011} designed a marker composed of small circles spread on the edge of two or more concentric rings.
The image of each circle is used with a vote system to estimate the focal length and the image of the external rings.

Two circles on a planar marker (except if one encloses the other) is the minimum to fully estimate the homography without any other assumptions.
However in some applications e.g., dealing with concentric circles, detecting the images of two or more circles can be tricky.
First because the lack of points induces an inaccurate estimation and, secondly because it is time consuming.
When the camera has already been calibrated, it is possible to compute the homography from one circle image with two ambiguities.
Pagani et al.~\cite{Pagani2011} introduced a method quite similar to the solution proposed by Chen et al.~\cite{Chen2004a}, where the
ambiguity is solved by minimizing a distance between the image of the marker rectified and the expected pattern on all possible poses.

\section{Pose estimation from the image of one circle}
\label{sec:reminder}

We remind here some geometrical background on the problem of
  pose estimation from the image of  a single circle.
We consider an Euclidean projective camera, represented
by a  $3\times 4\text{-matrix}$
$\mathsf{P} \sim \mathsf{K}\mathsf{R}
                                \begin{bmatrix} 
                                        \mathsf{I} \mid \mathbf{T} \\
                                \end{bmatrix}$,
where the rotation matrix 
$\mathsf{R} \in SO(3)$\footnote{$SO(3)$ refers to the  3D rotation group.}
and the translation vector $\mathbf{T}\in  \mathbb{R}^3$  describe  the pose
of the camera, i.e., respectively its orientation
and  position in the object 3D frame.
The upper triangular order-$3$ matrix $\mathsf{K}$ is the  calibration matrix
as defined in~\cite[page 157]{RichardHartley2003}.
%
%

Assume that  $\mathcal{P}$ is a plane with equation $Z=0$ in the world
frame. The pose of  $\mathcal{P}$ in the camera frame is given by the vector
$[\mathbf{N}=\mathbf{r}_3,-d]^\top$, where  $\mathbf{r}_3$,  the
third column of $\mathsf{R}$,
defines the unit norm $\mathbf{N}$ of
  $\mathcal{P}$, and $d$ is the orthogonal distance to
  $\mathcal{P}$. The restriction to  $\mathcal{P}$
 of the projection mapping is an homography whose matrix writes
        $ \mathsf{H} \sim 
        \mathsf{K}\mathsf{R}
        \begin{bmatrix} 
                \mathbf{e}_1 \mid \mathbf{e}_2 \mid \mathbf{T} \\
        \end{bmatrix} 
        $, where $\mathbf{e}_1$ and $\mathbf{e}_2$ are the first two  columns
of $\mathsf{I}$.
In the projective plane, any conic can be represented in 2D homogeneous coordinates
by a real symmetric   order-$3$
matrix.
Under perspective projection, any circle of   $\mathcal{P}$, assuming  its
quasi-affine invariance \cite[p515]{RichardHartley2003} i.e., that all its points lie in front of the camera,
 is mapped  under
the homography $\mathsf{H}$ to an ellipse by the projection equation
$\mathsf{C}= \mathsf{H}^{-\top} \mathsf{Q} \mathsf{H}^{\top}$, where   
$\mathsf{Q}\in\text{Sym}_3$\footnote{$\text{Sym}_3$ refers to the space
of order-3 real symmetric matrices.} is the circle matrix and  $\mathsf{C}\in\text{Sym}_3$
is
the
ellipse matrix.

For reasons that will become clearer later, we want to parameterize the homography
 $\mathsf{H}$,  \textit{from only the
knowledge of the circle  image $\mathsf{C}$ and  the vanishing line $\mathbf{v}_\infty$
of 
 $\mathcal{P}$}.
Let   $\mathsf{S}_\mathcal{P}$ be a similarity
on the world plane that puts the circle $\mathsf{Q}$ into a unit circle centered
at
the origin and   $\mathsf{S}_\mathcal{I}$ be a similarity on the image plane
$\mathcal{P}$ that puts  $\mathsf{C}$ into a canonical
diagonal form $\mathsf{C}'=\operatorname*{diag}(C'_{11},C'_{22},C'_{33})$.
Using an approach similar to~\cite{Calvet2016} with the notation  $[u,v,1]^\top\sim\mathsf{C}^{-1}\mathbf{v}_\infty$,
{it can be shown that, under the assumption of a camera
with square pixels, we have  $\mathsf{H} \sim\mathsf{S}_\mathcal{I}^{-1}\mathsf{M}\mathsf{S}_\mathcal{P}$
}where 
\begin{align}\label{equ:050}
\mathsf{M}&=
\begin{bmatrix}
-1 & C'_{22}uv & -u \\
0 & - C'_{11}u^2 +1 & -v \\
-C'_{11}u & C'_{22}v & 1 \\
\end{bmatrix}
\begin{bmatrix}
r & 0 & 0 \\
0 & -1 & 0 \\
0 & 0 & s \\
\end{bmatrix}\notag\\
\text{ with }
r&=(-\frac{C'_{22}}{C'_{11}}(C'_{11}u^2 + C'_{22}v^2 + C'_{33}))^{1/2}\\
\text{ and }s&=\left(-C'_{22}(1-C'_{11}u^2 )\right)^{1/2}\notag
\end{align}
Note that the matrices   $\mathsf{S}_\mathcal{P}$  and    $\mathsf{S}_\mathcal{I}$
can be completely determined by the circle image  $\mathsf{C}$  and $\mathsf{M}$,
except for an unknown 2D rotation around the circle centre on $\mathcal{P}$.
  Recovering this rotation is not the goal of this
paper.
Some simple solution like placing a visible mark on the edge of the marker works generally well in many cases.

Our main task will be to recover the vanishing line $\mathbf{v}_\infty$ of
the plane, as explained in the sequel.
Note that the vector $\mathbf{x}_c=[u,v,1]^\top$ defined above is
that of the image of the circle centre which is the pole of $\mathbf{v}_\infty$
w.r.t. the dual ellipse of $\mathsf{C}$.

\section{Support plane's vanishing line estimation}
\label{sec:vanish_line}

We warn the reader that parts written in italics in  this section requires a proof that is not provided due to lack of space. However all proofs will appear in an extended paper version.   
\subsection{A twofold solution in the calibrated case}
\label{sec:calib_case}

In the case of calibrated image, 
an equivalent problem of computing the pose of its support plane $\mathcal{P}$ is that of recovering
the vanishing line of $\mathcal{P}$. 
Let $\mathsf{Q}$ be the matrix of a circle on a plane $\mathcal{P}$, and
$\bm\psi=\mathsf{H}^{\top}\bm\omega\mathsf{H}$ be that of the
back-projection onto $\mathcal{P}$ of the image  of the absolute conic~\cite[p.
81]{RichardHartley2003},
where $\bm\omega=\mathsf{K}^{-\top}\mathsf{K}^{-1}$. \textit{It is easy to
show that $\bm\psi$   represents  also a virtual\footnote{Virtual
conics have  positive definite matrices, so, no real points on them.  } circle
(as does $\bm\omega$)}.

Let $\left\{\alpha_{i}\right\}_{i=1..3}$ denotes the set of generalized eigenvalues
of the matrix-pair $(\mathsf{Q}, \bm\psi  )$, i.e., the three roots of the
characteristic equation $\det(\mathsf{Q}-\alpha\bm\psi)=0$. 
The set of matrices $\left\{\mathsf{Q}-\alpha\bm\psi\right\}_{\alpha\in \mathbb{R}\cup\left\{
\infty \right\}}$ defines a conic pencil~\cite{Gurdjos2006} which includes
three degenerate conics with matrices $\mathsf{D}_i~=~\mathsf{Q}-\alpha_{i}\bm\psi$.
These rank-2 matrices represent line-pairs and have the form  
$\mathsf{D}_i~=~\mathbf{l}_{a}^i(\mathbf{l}_{b}^i)^\top+\mathbf{l}_{b}^i(\mathbf{l}_{a}^i)^\top$,
where $\mathbf{l}_{a}^i$ and $\mathbf{l}_{b}^i$ are  vectors of these lines.
Such line-pair matrix  $\mathsf{D}_i$ can be easily decomposed and vectors
of its lines recovered albeit it is impossible to distinguish  $\mathbf{l}_{a}^i$
from  $\mathbf{l}_{b}^i$.
\textit{It can be shown that the  projective signatures\footnote{The signature
of a conic
is $\sign{\mathsf{C}}=(\max(p,n),\min(p,n))$, where  $p$ and $n$ 
count the positive and negative eigenvalues of its (real) matrix $\mathsf{C}$.
It is left unchanged by projective transformations.}
of the three degenerate members always are $(1,1)$, $(2,0)$ and $(0,2)$}.
Assume, without loss of generality, that the degenerate conic $\mathsf{D}_{2}$
is the one with signature $(1,1)$.
\textbf{\textit{A first key result}} is that  $\mathsf{D}_{2}$ is a pair
of  two distinct real lines, one of which being  
 the {line at infinity} $\mathbf{l}_\infty=[0,0,1]^\top$; the
other one being denoted by $\mathbf{l_o}$.
The other  two degenerate conics $\mathsf{D}_{1}$ and $\mathsf{D}_{3}$  ---with
signatures $(2,0)$ and $(0,2)$---  are
pairs of two {conjugate complex lines}.
Consequently, the three  (so-called) {base points} $\mathbf{x}_i$,
where lines in a pair meet, are real. Moreover, 
their vectors are the  generalized eigenvectors of $(\mathsf{Q}, \bm\psi
 )$ and satisfy $\mathsf{D}_i\mathbf{x}_i=\mathsf{0}$.

Similarly, in the image plane, if  $\mathsf{C}$  denotes the image of the
circle  $\mathsf{Q}$,
the set of matrices $\left\{\mathsf{C}-\beta\bm\omega\right\}_{\beta\in\mathbb{R}\cup\left\{
\infty \right\}}$ defines also a conic pencil whose members are the images
of the pencil 
 $\left\{\mathsf{Q}-\alpha\bm\psi\right\}_{\alpha\in\mathbb{R}\cup\left\{
\infty \right\}}$.
Hence, the line-pair in $\left\{\mathsf{C}-\beta\bm\omega\right\}$ that includes
the
image of  $\mathbf{l}_\infty$
i.e., the vanishing line $\mathbf{v}_\infty$,  can always be identified since
it is the only degenerate member  with signature $(1,1)$.
Nevertheless, at this step, it is impossible to distinguish  $\mathbf{v}_\infty$
from the other line   $\mathbf{v_o}$, image of   $\mathbf{l_o}$. 

Assume that
 all matrices $\mathsf{Q}$, $\mathsf{C}$, $\bm\psi$ and $\bm\omega$ 
are normalized
to have a unit determinant. It is known that, in this case, parameters
in pencils satisfy $\alpha~=~\beta $, so, the generalized
eigenvalues of the matrix-pair $(\mathsf{Q},\bm\psi)$
 are \textit{exactly} the same
as those of $(\mathsf{C}, \bm\omega  )$. 
\textit{It can be shown that these eigenvalues can always be sorted such
that $\lambda_1\ge\lambda_2\ge\lambda_3$}, where 
$\mathsf{D_2}~=~\mathsf{C}-\lambda_2 \bm\omega$ is the (sole) degenerate
conic with signature $(1,1)$.
Remind that $\mathsf{D_2} $ is the conic which contains $\mathbf{v}_\infty$
plus $\mathbf{v_o}$, which are two \textit{a priori} indistinguishable
lines  denoted by $\mathbf{v_{1,2}}$.
Because the matrix $\mathsf{D_2} $ is real, symmetric, rank-$2$ and order-$3$,
\textit{its generalized eigen-decomposition using  the   
 base point vectors 
$\mathbf{  x}_1,\mathbf{  x}_3\in\mathbb{R}^3$  writes as following:
}\begin{equation}
        \mathsf{D_2} = \begin{bmatrix} \frac{\mathbf{  x_1}}{\left\Vert\mathbf{
 x_1}\right\Vert} & \frac{\mathbf{  x_3}}{\left\Vert\mathbf{  x_3}\right\Vert}
\end{bmatrix} 
\begin{bmatrix} 
\lambda_1-\lambda_2 & 0 
\\ 0 & \lambda_3-\lambda_2 
\end{bmatrix} 
\begin{bmatrix} 
\frac{\mathbf{  x_1^{\top}}}{\left\Vert\mathbf{  x_1}\right\Vert}  \\ 
\frac{\mathbf{  x_3^{\top}}}{\left\Vert\mathbf{  x_3}\right\Vert}
\end{bmatrix}
\end{equation}
from which \textit{it can be shown that} 
\begin{equation}\label{equ:xxx:001}
        \mathbf{v_{1,2}} = \sqrt{\lambda_1-\lambda_2} \frac{\mathbf{  x_1}}{\left\Vert\mathbf{
 x_1}\right\Vert} \pm
\sqrt{\lambda_2-\lambda_3}\frac{\mathbf{  x_3}}{\left\Vert\mathbf{
 x_3}\right\Vert}
\end{equation}
The two solutions to the normal to $\mathcal{P}$ are given by $\mathbf{N_{1,2}}~=~\mathsf{K}^{\top}\mathbf{v_{1,2}}$
in the camera frame,
and (\ref{equ:xxx:001}) explains the known doublefold ambiguity in the plane
pose~\cite{Chen2004a}. 

\subsection{About removing the twofold ambiguity}

\begin{figure}[h]
	\includegraphics[scale=0.3]{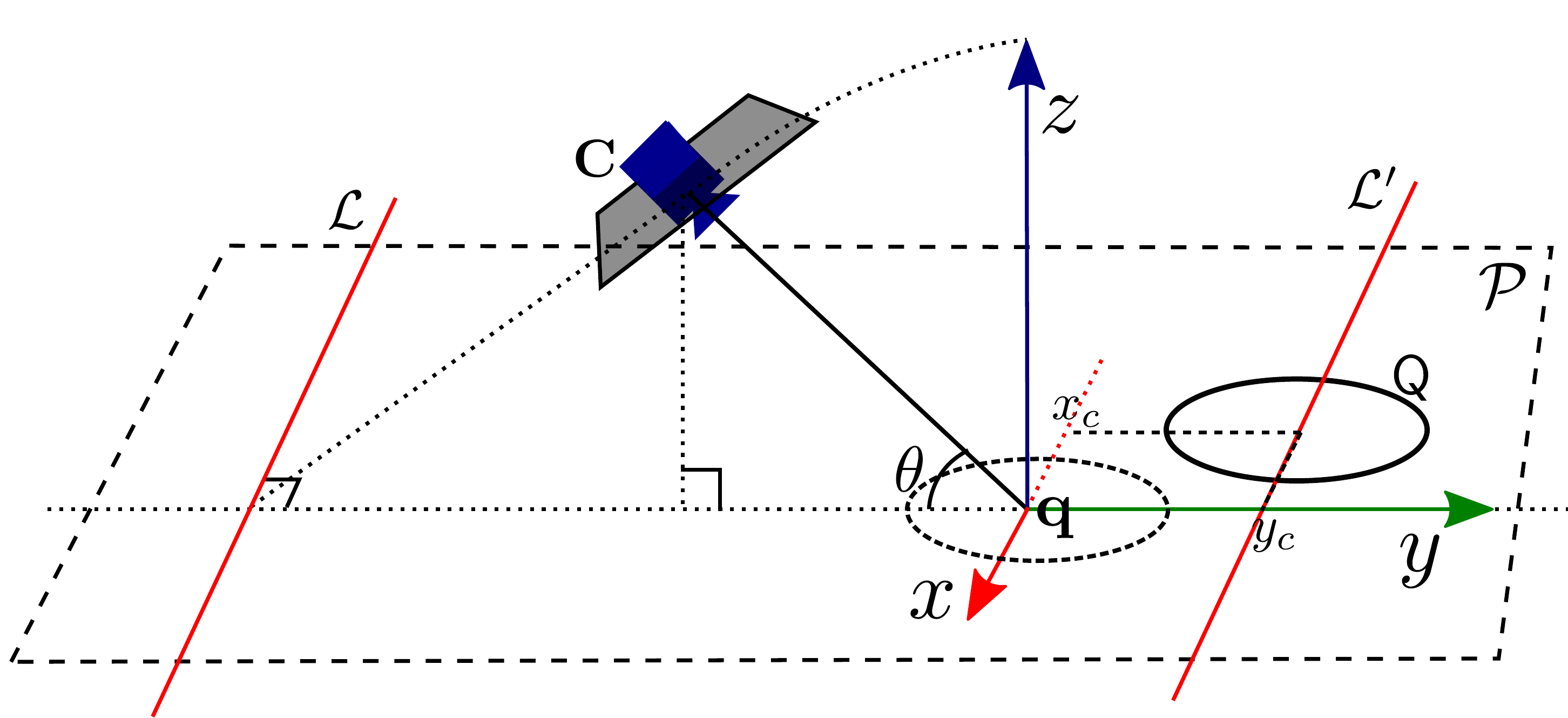}
    \caption{Proposed parametrization for detecting the ambiguity.}
    \label{fig:ambig}
\end{figure}

We have seen that there are two solutions for the vanishing line (or the
plane normal in the calibrated case) which are in general not distinguishable.
In this section, we  discuss whether known configurations allows the ambiguity
to be removed. We extend the new theoretical framework
proposed in \S\ref{sec:calib_case} that involves the point $\mathbf{q}$ (on
the support
plane $\mathcal{P}$) where
the optical axis cuts $\mathcal{P}$ plus the line $\mathcal{L}$ obtained
by intersecting  $\mathcal{P}$ and the principal plane\footnote{The 3D plane
through the camera centre and parallel to the image
plane.} of the camera (${\mathcal{L}}$ is orthogonal to the orthogonal projection
of
the optical axis onto $\mathcal{P}$). Now, let ${\mathcal{L}'}$ denote
the line parallel
to $\mathcal{L}$ through the
circle centre.
Within this geometrical framework, we can claim, for instance,  that \textit{a sufficient condition for the
ambiguity to be solved is given by the two following conditions:
\begin{enumerate}
\item[\emph{(i)}] $\mathbf{q}$ and the orthogonal projection on $\mathcal{P}$
of the camera centre  lie on the same side of ${\mathcal{L}'}$ ;
\item[\emph{(ii)}] the point, intersection of the orthogonal projection on $\mathcal{P}$ of the optical axis and ${\mathcal{L}'}$, lies outside the circle centered at $\mathbf{q}$ with same radius as $\mathsf{Q}$. 
\end{enumerate}
} 
Figure \ref{fig:ambig} illustrates this important result.
We are convinced that future investigations using this framework can help
to reveal more  configurations in which the  ambiguity
can be removed.
We are now giving more geometrical insights indciating how to determine such configurations, via three propositions.   
The first is the \textbf{\textit{second key result}} which is the building brick
of our approach:

\vspace*{-3mm}
\begin{prp}[second key result]\label{prop:001}
The line $\mathbf{l_o}$
in  $\mathsf{D}_{2}$ 
separates the two base points  
  $\mathbf{x_1}$ and $\mathbf{x_3}$. Hence, denoting by $\mathbf{\bar x}$
the normalized vector $\mathbf{\bar x}=\mathbf{x}/x_3$, the  following inequalities
hold $ (\mathbf{l}_\infty^\top\mathbf{\bar x}_1)(\mathbf{l}_\infty^\top\mathbf{\bar
x}_3)>0$ and
$(\mathbf{l_o}^\top\mathbf{\bar x}_1)(\mathbf{l_o}^\top\mathbf{\bar x}_3)<0$.


\end{prp}
These
two inequalities hold under any affine transformation but not under a general
projective transformation.

How the conditions in proposition \ref{prop:001} can be helpful in removing
the plane
pose ambiguity?
Can we
state a corollary saying that, in the
image plane, under some known geometric configuration, we know which the line $\mathbf{v_o}$
in $\mathsf{C}-\lambda_{2}\bm\omega$,
image of $\mathbf{l_o}$, 
\textit{always} separates points  
  $\mathbf{z_1}$ and $\mathbf{z_3}$,  images  of  base points   
  $\mathbf{x_1}$ and $\mathbf{x_3}$, while the other does not?
%
That is,
if we a priori know\ $\operatorname*{sign}(\mathbf{v}_o^\top\mathbf{\bar z_1})(\mathbf{v}_o^\top\mathbf{\bar
z_3})$ can we  guarantee that $(\mathbf{v}_o^\top\mathbf{\bar z_1})(\mathbf{v}_o^\top\mathbf{\bar
z_3})=-(\mathbf{v}_\infty^\top\mathbf{\bar z_1})(\mathbf{v}_\infty^\top\mathbf{\bar
z_3})$?
If yes, since the vectors of these base points are the generalized eigenvectors
of $(\mathsf{C},\bm\omega)$
associated to parameters $\lambda_j$, $j\in\left\{ 1,3 \right\}$ and can
be straightforwardly computed,
 we could remove the ambiguity by choosing as vanishing line $\mathbf{v}_\infty$
the ``correct''  line in $\mathsf{C}-\lambda_{2}\bm\omega$.
%
We claim the following proposition  for
this corollary to hold, whose (omitted) proof directly
follows from the properties of  
quasi-affineness w.r.t. the base points~\cite{RichardHartley2003}.

\vspace*{-3mm}
\begin{prp}\label{prop:002}
When  $\mathbf{x_1}$ and $\mathbf{x_3}$  lie either both in front or   both
behind the camera i.e.,  on the same half-plane bounded by $\mathcal{L}$ , we have
  $(\mathbf{v}_o^\top\mathbf{\bar
 z_1})(\mathbf{v}_o^\top\mathbf{\bar
 z_3})<0$ and  $(\mathbf{v}_\infty^\top\mathbf{\bar z_1})(\mathbf{v}_\infty^\top\mathbf{\bar
 z_3})>0$. Otherwise 
  $(\mathbf{v}_o^\top\mathbf{\bar
 z_1})(\mathbf{v}_o^\top\mathbf{\bar
 z_3})>0$ and  $(\mathbf{v}_\infty^\top\mathbf{\bar z_1})(\mathbf{v}_\infty^\top\mathbf{\bar
 z_3})<0$. 
\end{prp}

Now let us investigate a formal condition saying  when  $\mathbf{x_1}$ and $\mathbf{x_3}$
 lie on the same half-plane bounded by $\mathcal{L}$.
Consider an Euclidean representation of the projective world in which the
origin is the point $\mathbf{q}$  at which the
optical axis cuts the plane $\mathcal{P}$. 
Let the  $X$-axis be parallel to the line $\mathcal{L}$ and the $Y$-axis is
 the orthogonal projection of
the optical axis onto $\mathcal{P}$.
Consequently, the $Z$-axis is directed by the normal to 
 $\mathcal{P}$, as shown in figure \ref{fig:ambig}.
Let $\mathbf{C}~=~[0,-\cos\theta,\sin\theta]^\top$, 
$\theta\in[0,\frac{\pi}{2}[$,
be the 3D cartesian coordinates of the camera centre, where $\pi-\theta$
is the angle between the $Y$-axis and the optical axis
in the $YZ$-plane (note that we choose the scale such that the camera centre
is at distance $1$ from the origin).
Therefore the direction of the optical
axis is given by $-\mathbf{C}$.
 
\vspace*{-3mm}

In the 2D  representation of the projective plane  $\mathcal{P}$ 
 (i.e., of the $XZ$-plane), let the circle have centre $(x_c,y_c)$ and radius
$R$. 
Let  $\mathbf{d}=[0,1,\cos\theta]^\top$ is the  vector of line
$\mathcal{L}$. It can be shown, using a symbolic software like \textsc{maple}\footnote{https://fr.maplesoft.com/},
that:

\vspace*{-3mm}
\begin{prp}
Base points $\mathbf{x_1}$ and $\mathbf{x_3}$
lie,
in the world plane, on the same side of $\mathcal{L}$ \textit{if and only if} 
\begin{equation}\label{equ:18:01:12:01}
cos\theta(y_c^2-R^2)(yc + cos\theta) 
 + y_ccos\theta (1+ x_c^2)+ x_c^2 + y_c^2 \le 0
\end{equation}  
\end{prp}
Since $cos\theta>0$, if   $yc >0$ and $y_c^2-R^2 > 0$  
then $\mathbf{x_1}$ and $\mathbf{x_3}$
lie on opposite sides of $\mathcal{L}$.
The former inequality says that $\mathbf{q}$ must lie on the same side of $\mathcal{L}'$, the line parallel
to $\mathcal{L}$ through the
circle centre, as the orthogonal projection
of the camera centre  onto $\mathcal{P}$.  The latter inequality  says the point $(0,y_c)$ must lie outside the circle centered at $\mathbf{q}(0,0)$ with same radius $R$ as $\mathsf{Q}$. 
As we are in the ``otherwise'' part of proposition \ref{prop:002}, 
the vanishing line is given by the line that does not separate the
image of the base points. Since $(0,y_c)$ represents the intersection of the orthogonal projection on $\mathcal{P}$ of the optical axis and ${\mathcal{L}'}$, this is  the result announced at the beginning of this section.

\subsection{Defining default intrinsics for the camera}

\label{sec:gen_cam}

In the previous sections we have seen that, providing that the camera intrinsics
are known, there is a twofold solution for the vanishing line.
Recovering accurate intrinsics of a camera requires generally a calibration
procedure.
In many applications,  the model of the camera can be simplified to reduce
the number of parameters.
A very common   model is that of a camera with square pixels and  principal point at the centre of the image plane. 
Consequently, the focal length is the sole unknown,
e.g., for self-calibration purposes~\cite{Pollefeys1999}. The focal length
value is sometimes  available through 
 EXIF data,  stored in digital images or video files,
through camera hardware on top level API (Android, iOS) or
 through data provided by manufacturer on websites.
Focal length, denoted  $f$, in pixels (what we need) can be obtained from this data if we
find the field of view in angle or the focal length equivalent in $35$mm.
However the focal length is very often given in millimetre without
the sensor size required to obtain the focal length in pixels.
 
We consider here the case where it is impossible to calibrate the camera
by none of the methods mentioned above. So how to do?
We propose to design
a generic camera model delivering default intrinsics (i.e., focal length)
and based on off-line calibration of several smartphone cameras. 
If a camera can generally take any focal length value, the optics and the
sensor of smartphones are constrained by the device size and the desired
field of view.
Why doing that?
We found out
that surprisingly enough,
that it is not necessary to have very accurate intrinsics to estimate the
vanishing line given the image of a single circle.
In fact, as shown in the experimental section~\ref{sec:experiment}, this estimation is very robust to intrinsics fluctuation.

\begin{figure}[htp]
    \centering
    \includegraphics[scale=0.4]{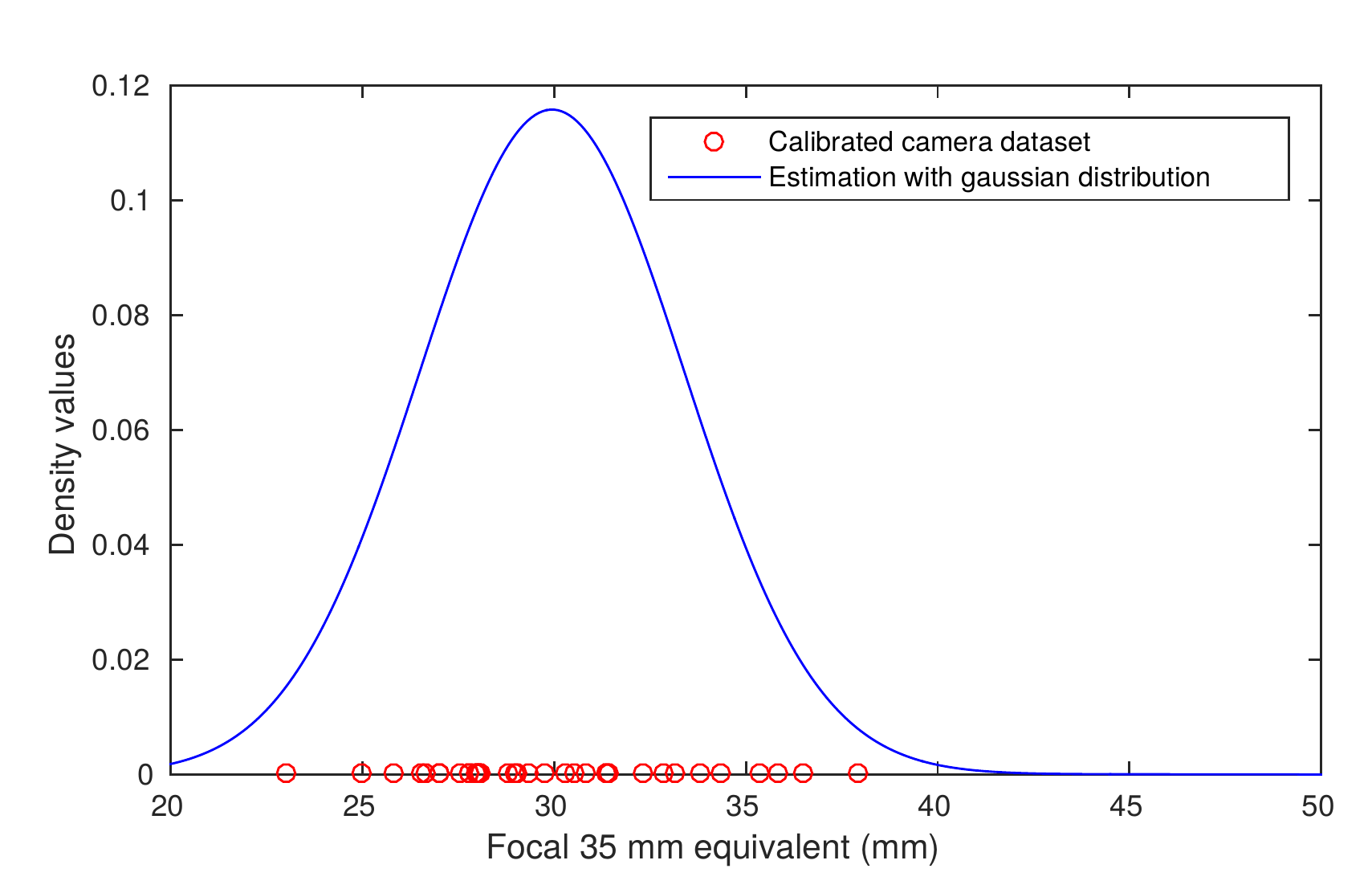}
    \caption{Focal calibration of different camera parameters}
    \label{fig:smart_calib}
\end{figure}

After calibrating a dozen of camera devices and obtaining data from manufacturers
of twenty more smartphones, we estimate a gaussian model of the focal length equivalent in 35mm, as shown in figure \ref{fig:smart_calib}.
In our case we obtained experimentally an average focal length of $f_{35}
= 29.95mm$ with a variance of $\sigma_{f35}^2 = 11.86$.
More precisely, we estimate a gaussian function (in blue) based of the focal
values collected or estimated (in red) from different smartphone device brands.

\section{Experimental results }
\label{sec:experiment}

\subsection{Test Description}

\begin{figure}[htp]
	\begin{tabular}{c l c r}
		\hspace{3pt} & \includegraphics[scale=0.45]{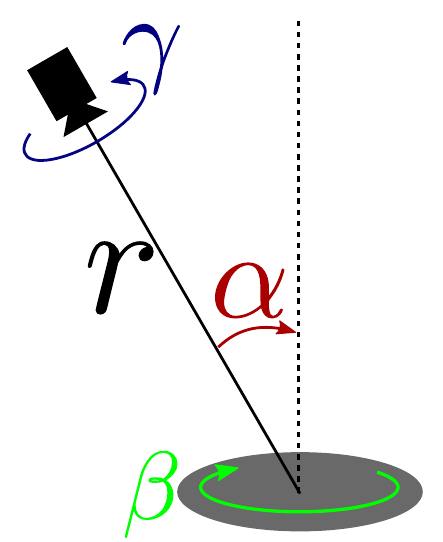} & \hspace{0.4cm} &
    	\includegraphics[scale=0.035]{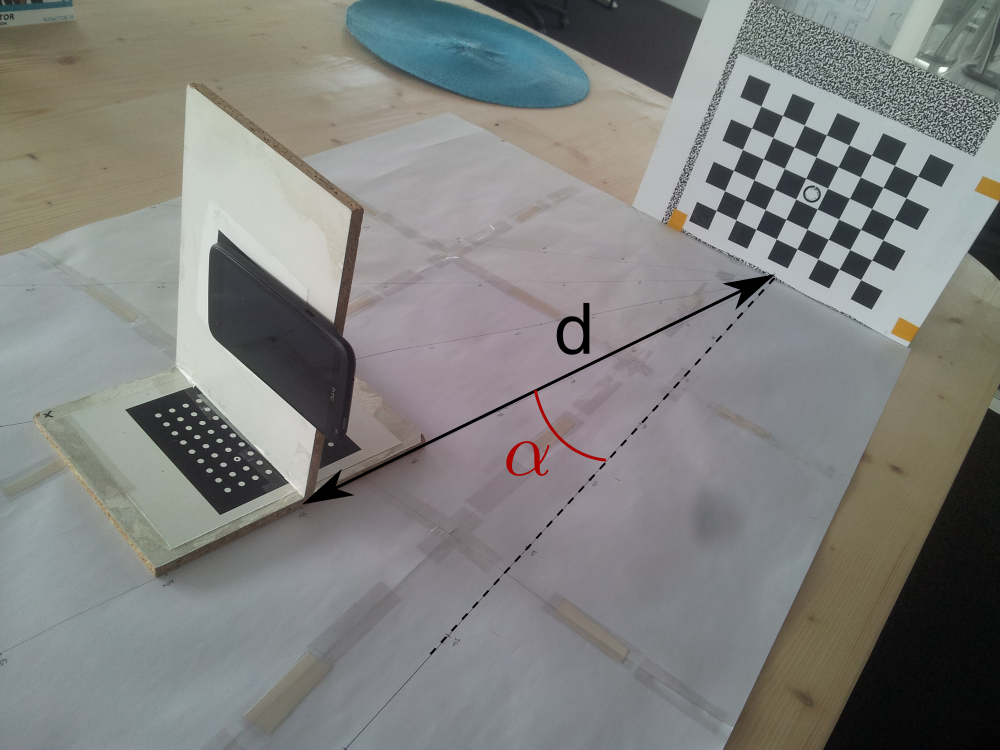} \\
    \end{tabular}
	\caption{Setup: reference chessboard and pose annotation}
	\label{fig:pose}
\end{figure}

The goal of the test presented in this section is to evaluate the proposed method to estimate the pose of a camera.
We performed those tests on synthetic and real images in the conditions illustrated in figure \ref{fig:pose}.
In order to limit the poses used in experiments, we made some hypotheses.
First, we suppose that the camera focus on the centre of the marker, i.e. the principal axis of the camera passes through the centre of the marker, see figure~\ref{fig:pose}.
Then, the angle $\gamma$ has been set to zero.
In deed, we can simulate any angle by rotating the image using the assumption that the principal point is centred on the image and that the principal axis is orthogonal to the image plane.
Finally, the angle $\beta$ has been fixed to zero as estimating the 2D rotation around the plane normal is out of the scope of this article.
The remaining variables whose variations are studied in our test are the angle $\alpha$ and the distance $r$.

We know that introducing generic camera parameter, as proposed in section~\ref{sec:gen_cam}, should have a negative impact on the accuracy of the pose estimation.
Consequently, one of the objectives of this experiment is to evaluate the sensitivity of the proposed method to inaccurate camera focal parameter.
The observation of the distribution of focal length of various smartphone camera, see figure~\ref{fig:smart_calib}, reveals that all 35mm focal equivalent are included in $[-30 \%, +30 \%]$ of the average value. 
So, five different values that span this range are used in the experiment: $\{0.7,0.85,1.0,1.15,1.3\}$.
In order to generate synthetic images, we have simulated a synthetic camera of focal $\alpha_x = 1280$ and resolution of $1280\times 720$ pixels.
To obtain real images, we have used the camera of a smartphone 
which have been calibrated with openCV library\footnote{https://opencv.org/}.
In both cases, we suppose that ellipses have been firstly detected in the images, i.e. contour points are first detected and then ellipses are estimated~\cite{Szpak2015}.
We try to evaluate the impact of errors of this estimation to the quality of the results.
In consequences, in our synthetic tests, we have also simulated noises on the detections of the ellipses, i.e. errors on the pixels that belong to the ellipse.
More precisely, edge points of the ellipse have been translated with a zero mean gaussian variance of $\sigma_x = 1.0$. 

Finally, we evaluate the quality of the results obtained by using three different measurements relative to the pose and the reprojection accuracy:
\begin{enumerate}[label=\alph*)]
	\item Error on the normal of the plane relative to the camera;
	\item Error on the position of the marker;
	\item Error of reprojection of 3D points close to the marker.
\end{enumerate}
Each curve illustrates the results obtained by applying a modifier on focal length used for pose estimation.
The resulting errors are displayed as function of the distance $r$ in the interval $[15\times D,50 \times D]$ where $D$ is the diameter of the marker. 
This interval is related to the distances used for 
being able to detect and to recognize a marker for an augmented reality application, i.e. the distance where the marker occupies, at least 80 pixels. 
We also show results for three different angle values, $\alpha \in \{15,30,45\}$, displayed in three sub-figures.

%
%

\subsection{Analysis of the results}
\begin{figure*}[htp]
	\begin{center}
	\subfloat[Orientation of the normal]{
	\includegraphics[scale=0.35]{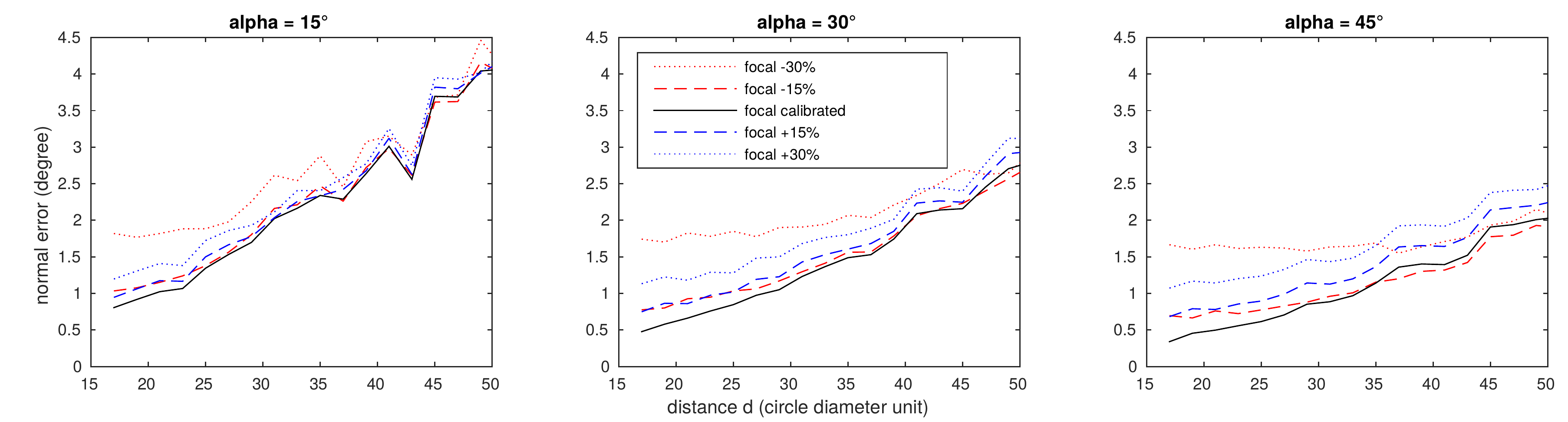}
        \label{fig:res_synth_normal}
    }
	\vspace{-0.24cm}
    \subfloat[Position of the marker]{
    	\includegraphics[scale=0.35]{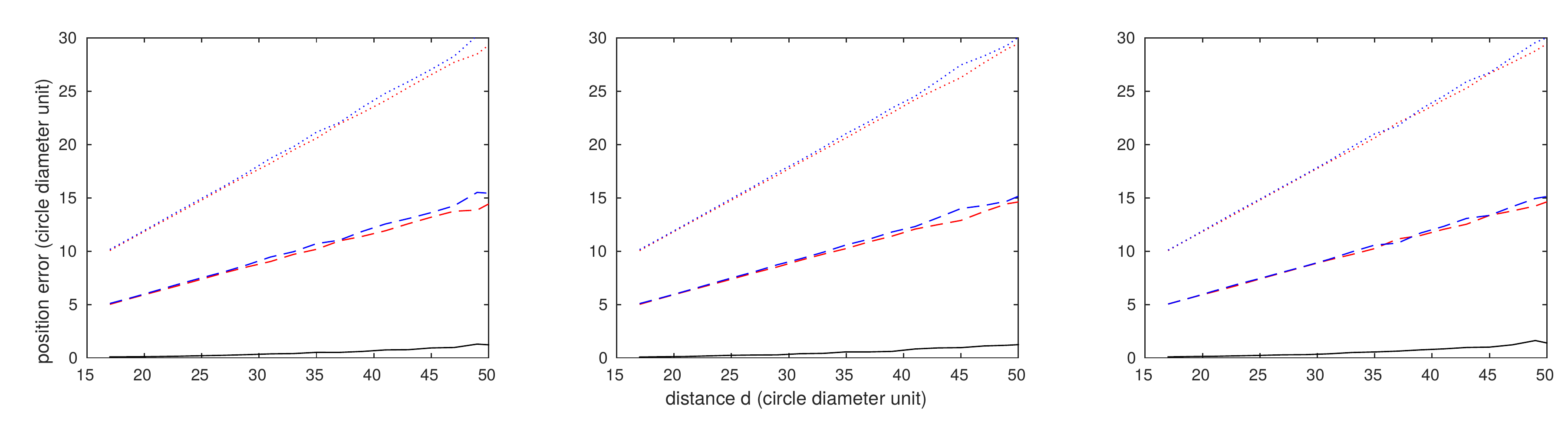}
        \label{fig:res_synth_pos}
    }
    \vspace{-0.24cm}
    \subfloat[Reprojection of 3D points]{
    	\includegraphics[scale=0.35]{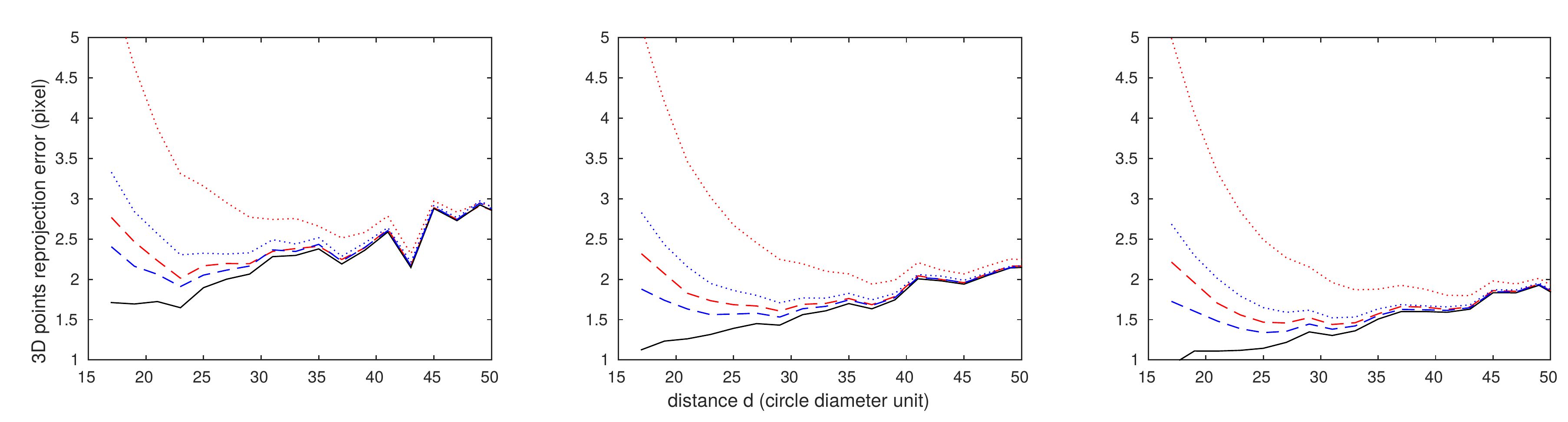}
        \label{fig:res_synth_3D}
    }
    \end{center}
    \caption{Error with synthetic images.}
    \label{fig:res_synth}
\end{figure*}

Results on synthetic images are presented in figure~\ref{fig:res_synth}. 
In~\ref{fig:res_synth_normal}, we show the error on the estimation of the orientation for the pose.
We can notice that as the distance of the marker to the camera increases, the error on pose orientation also increases.
This relation is even more remarkable when the angle is the lowest between the marker plane and the camera, i.e. the graph on the left.
In~\ref{fig:res_synth_pos}, we can see that in the calibrated case the accuracy in position stays low and does not depend on the distance to the camera and the angles between the marker plane and the camera.
In the uncalibrated cases, as expected the detection of the ellipses becomes less accurate when the distance increases and, consequently, the quality of the estimation of the marker position is also affected. In fact, the error in position increases linearly when the distance increases. 
This observation is quite intuitive. 
In deed observing a marker with a zoom or taking its image closer leads to very similar shape of the marker.
The error on the reprojection of 3D points, presented in~\ref{fig:res_synth_3D}, illustrates that, with a focal length well estimated, the higher the distance, the higher the errors. Whereas, when the focal length is not well estimated, the higher the distance, the lower the error and, more important, this error is quite near the error obtained when the focal length is correctly estimated. 
It means that using generic parameter is not affecting the quality of the reprojection in a context where the marker is far from the camera. 

\begin{figure*}
	\vspace{-0.28cm}
	\begin{center}
    	\includegraphics[scale=0.35]{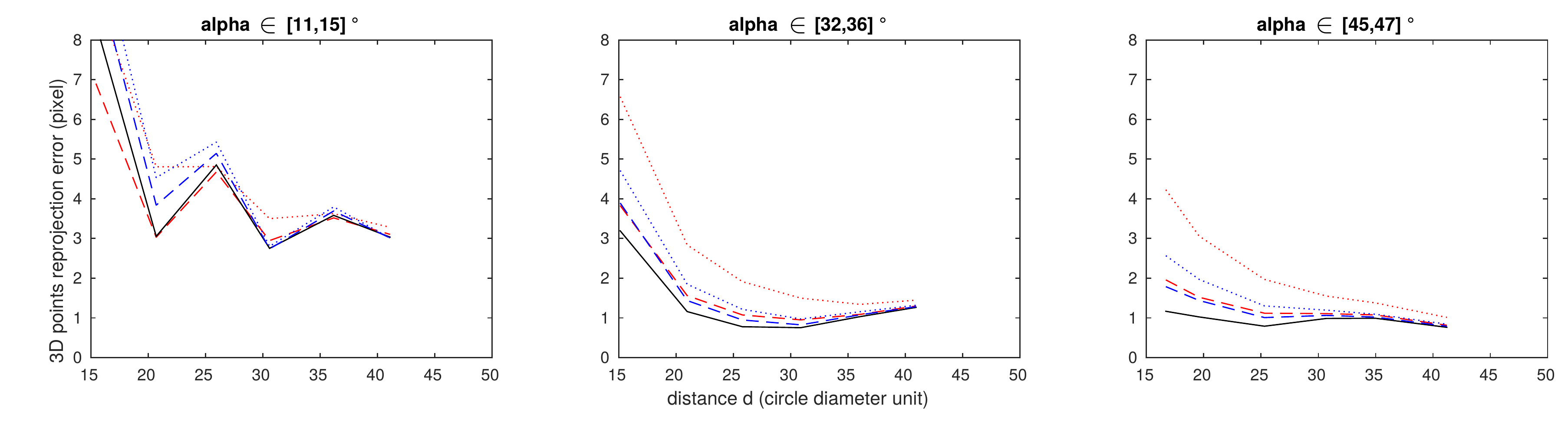}
    \end{center}
    \caption{Error with ellipses detected on real image: 3D points reprojection.}
    \label{fig:res_real}
\end{figure*}

The figure \ref{fig:res_real} allows us to present similar conclusions on real images. The 3D point reprojection error is 
presented. 
The error in calibrated case slightly increases with the distance as observed
 in figure~\ref{fig:res_synth_normal}.
When the marker is close to the camera, the error of reprojection when the camera is not correctly calibrated is high but it drastically decreases when the distance
to the camera increases, and, finally, this error is of the same order as that obtained with the calibrated case. 
This observation is not really a surprise as the projection of a distant object loses its perspective with distance.
Again, this result illustrates the interest of using generic camera parameter in augmented reality.

\section{Conclusion}

In this paper, we introduced a method to estimate the pose of a camera from the image of a circular marker in a calibrated case. If, in general case, two solutions are found, some assumptions on geometric configuration can help to distinguish the correct pose. Moreover, we demonstrated the interest of using default camera parameters, in the context of augmented reality. In particular, the results presented showed that, in a case of a distant marker, the 3D reprojection errors is low enough. 
Future work would be to use more information in the marker environment to increase the stability of the detection of the marker and the pose estimation and to allow decoding from longer distance. 

\ifCLASSOPTIONcompsoc
 \section*{Acknowledgments}
  We would like to thanks the company Ubleam, and R\'{e}gion Occitanie for the participation to the project INVISO which leads to this publication.
\else

%
%

\bibliography{biblio}

\begin{thebibliography}{10}

\bibitem{Bergamasco2011}
F.~Bergamasco, A.~Albarelli, E.~Rodol{\`{a}}, and A.~Torsello.
\newblock {RUNE-Tag: A high accuracy fiducial marker with strong occlusion
  resilience}.
\newblock {\em CVPR}, 2011.

\bibitem{Calvet2016}
L.~Calvet, P.~Gurdjos, C.~Griwodz, and S.~Gasparini.
\newblock {Detection and Accurate Localization of Circular Fiducials under
  Highly Challenging Conditions}.
\newblock {\em CVPR}, 2016.

\bibitem{Chen2004a}
Q.~Chen, H.~Wu, and T.~Wada.
\newblock {Camera calibration with two arbitrary coplanar circles}.
\newblock {\em ECCV}, 2004.

\bibitem{Fiala2010}
M.~Fiala.
\newblock {Designing highly reliable fiducial markers}.
\newblock {\em PAMI}, 2010.

\bibitem{Gurdjos2006}
P.~Gurdjos, P.~Sturm, and Y.~Wu.
\newblock {Euclidean structure from N {\textgreater}= 2 parallel circles:
  Theory and algorithms}.
\newblock {\em ECCV}, 2006.

\bibitem{RichardHartley2003}
R.~Hartley and A.~Zisserman.
\newblock {\em {Multiple View Geometry}}.
\newblock 2003.

\bibitem{Huang2015}
H.~Huang, H.~Zhang, and Y.~M. Cheung.
\newblock {The common self-polar triangle of concentric circles and its
  application to camera calibration}.
\newblock {\em CVPR}, 2015.

\bibitem{Kim2005}
J.~S. Kim, P.~Gurdjos, and I.~S. Kweon.
\newblock {Geometric and algebraic constraints of projected concentric circles
  and their applications to camera calibration}.
\newblock {\em PAMI}, 2005.

\bibitem{Pagani2011}
A.~Pagani, J.~Koehle, and D.~Stricker.
\newblock {Circular Markers for camera pose estimation}.
\newblock {\em Image Analysis for Multimedia Interactive Services}, 2011.

\bibitem{Pollefeys1999}
M~Pollefeys, R~Koch, and L~Van Gool.
\newblock {Self-Calibration and Metric Reconstruction in Spite of Varying and
  Unkown Internal Camera Parameters}.
\newblock {\em IJCV}, 1999.

\bibitem{Szeliski2011}
R.~Szeliski.
\newblock {\em Computer Vision: Algorithms and Applications}.
\newblock 2010.

\bibitem{Szpak2015}
Z.~L. Szpak, W.~Chojnacki, and A.~van~den Hengel.
\newblock {Guaranteed Ellipse Fitting with a Confidence Region and an
  Uncertainty Measure for Centre, Axes, and Orientation}.
\newblock {\em JMIV}, 2015.

\bibitem{Wu2004}
H.~Wu, Q.~Chen, and T.~Wada.
\newblock {Conic-based algorithm for visual line estimation from one image}.
\newblock {\em Automatic Face and Gesture Recognition}, 2004.

\end{thebibliography}
\bibliographystyle{plain}

%
%

\end{document}